\documentclass{article}

\usepackage[final]{neurips_2024}

\usepackage[utf8]{inputenc} 
\usepackage[T1]{fontenc}    
\usepackage{hyperref}       
\usepackage{url}            
\usepackage{booktabs}       
\usepackage{amsfonts}       
\usepackage{nicefrac}       
\usepackage{microtype}      
\usepackage{xcolor}         
\usepackage{float}
\usepackage{subfig}
\usepackage{graphicx}
\usepackage{amssymb}
\usepackage{amsmath}
\usepackage{amsthm}
\usepackage{multirow}
\usepackage{adjustbox}
\usepackage[linesnumbered,ruled,vlined]{algorithm2e}
\usepackage{makecell}
\usepackage{bm}
\usepackage{caption}
\usepackage{natbib}
\setcitestyle{numbers}
\setcitestyle{square}
\setcitestyle{comma} 
\usepackage{enumitem}
\usepackage{bbding}
\usepackage{wrapfig}
\usepackage{mathtools}
\usepackage[toc,page,header]{appendix}
\usepackage{sidecap}

\usepackage{tabularx}
\setlist[itemize]{leftmargin=*}

\usepackage{etoc}
\etocdepthtag.toc{mtchapter}
\etocsettagdepth{mtchapter}{subsection}
\etocsettagdepth{mtappendix}{none}

\usepackage{colortbl,color}
\usepackage{hyperref}

\definecolor{citeColor}{RGB}{0,20,115}
\hypersetup{colorlinks,linkcolor={citeColor},citecolor={citeColor},urlcolor={citeColor}}  

\usepackage[textsize=tiny]{todonotes}

\newcolumntype{L}[1]{>{\raggedright\let\newline\\\arraybackslash\hspace{0pt}}m{#1}}
\newcolumntype{C}[1]{>{\centering\let\newline  \\\arraybackslash\hspace{0pt}}m{#1}}%
\newcolumntype{R}[1]{>{\raggedleft\let\newline \\\arraybackslash\hspace{0pt}}m{#1}}

\def\bW{\textbf{W}}
\def\bal{\bm{\alpha}}

\newtheorem{definition}{Definition}
\DeclareMathOperator*{\argmin}{arg\,min}
\DeclareMathOperator*{\argmax}{arg\,max}
\newcommand{\std}[1]{\scriptsize{$\pm$#1}}

\title{Customized Subgraph Selection and Encoding for Drug-drug Interaction Prediction}

\makeatletter
\renewcommand*{\@fnsymbol}[1]{\ensuremath{\ifcase#1\or \dagger\or \ddagger\or
		\mathsection\or \mathparagraph\or \|\or **\or \dagger\dagger
		\or \ddagger\ddagger \else\@ctrerr\fi}}
\makeatother

\author{%
	\textbf{Haotong Du}$^{1}$ \quad
	\textbf{Quanming Yao}$^{2}$ \quad
	\textbf{Juzheng Zhang}$^{2}$ \quad
	\textbf{Yang Liu}$^{1}$ \quad
	\textbf{Zhen Wang}$^{1}\thanks{
		Corresponding author.}$ 
	\\
	$^{1}$Northwestern Polytechnical University \quad
	$^{2}$Tsinghua University  \quad
	\\
	\texttt{duhaotong@mail.nwpu.edu.cn} \quad
	\texttt{w-zhen@nwpu.edu.cn} \\
	\texttt{qyaoaa@tsinghua.edu.cn} \quad
	\texttt{\{juzhengzh00,yangliuyh\}@gmail.com}
}

\begin{document}

\maketitle

\vspace{-8pt}
\begin{abstract}
	
Subgraph-based methods have proven to be effective and interpretable in predicting drug-drug interactions (DDIs),
which are essential for medical practice and drug development. 
Subgraph selection and encoding are critical stages in these methods, 
yet customizing these components remains underexplored due to the high cost of manual adjustments. 
In this study, 
inspired by the success of neural architecture search (NAS), 
we propose a method to search for data-specific components within subgraph-based frameworks. 
Specifically, 
we introduce extensive subgraph selection and encoding spaces that account for the diverse contexts of drug interactions in DDI prediction. 
To address the challenge of large search spaces and high sampling costs, 
we design a relaxation mechanism that uses an approximation strategy to efficiently explore optimal subgraph configurations. This approach allows for robust exploration of the search space. 
Extensive experiments demonstrate the effectiveness and superiority of the proposed method, 
with the discovered subgraphs and encoding functions highlighting the model’s adaptability.

\end{abstract}
\vspace{-10pt}
\section{Introduction}

Precise prediction of drug-drug interactions (DDIs) is essential in biomedicine and healthcare research~\cite{lin2023comprehensive}.
Drug combination therapy~\cite{mottonen1999comparison} can enhance treatment effectiveness for certain diseases; 
however, it also increases the risk of adverse drug reactions, 
potentially threatening patient safety~\cite{juurlink2003drug}.
Identifying DDIs through laboratory experiments is both costly and time-consuming~\cite{percha2013informatics,jiang2022adverse}.
With the success of deep learning, 
researchers have increasingly explored computational methods for DDI prediction. 
Early approaches primarily relied on molecular fingerprint information~\cite{rogers2010extended} or hand-engineered features~\cite{vilar2014similarity}, 
often neglecting the pre-existing interaction properties between drugs.

\begin{figure}[t]
	\centering
	\includegraphics[width=0.96\linewidth]{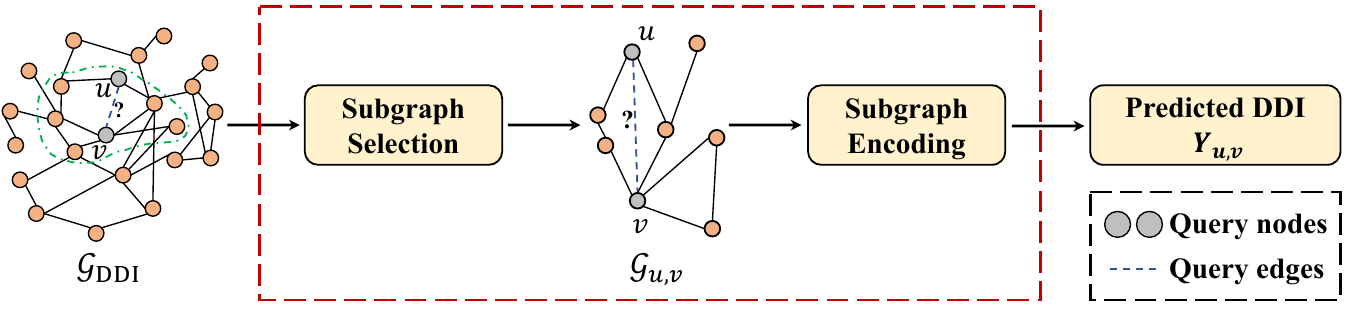}
	\caption{
		The pipeline of subgraph-based methods includes subgraph selection and subgraph encoding. 
		In this work, we focus specifically on searching for components within the red-dotted lines. 
	}	
	\label{fig-pipeline}
	\vspace{-18.5px}
\end{figure}

Considering drugs as nodes and their interactions as edges, 
DDI prediction can be framed as a multi-relational link prediction problem within the constructed drug interaction network.  
Recent advancements in graph neural networks (GNNs)~\cite{zitnik2018modeling,huang2020skipgnn,al2022prediction,le2023predicting} have consistently achieved superior
performance in this task.
Specifically, 
subgraph-based methods, 
such as SumGNN~\cite{yu2021sumgnn}, 
EmerGNN~\cite{zhang2023emerging}, 
and KnowDDI~\cite{wang2024accurate}, 
have shown promising results by selecting subgraphs around query edges and applying sophisticated encoding functions (message passing functions) to represent these subgraphs, 
Such methods transform the multi-relational link prediction task into a multi-type subgraph classification problem.
Figure~\ref{fig-pipeline} illustrates the pipeline of subgraph-based methods.

However, 
due to the dense nature~\cite{harada2020dual,udrescu2023curse} of drug interaction networks and their complex interaction semantics~\cite{nyamabo2022drug}, 
existing hand-designed subgraph methods often fail to capture the nuanced but crucial information across different data inputs.
In the initial phase of the reasoning pipeline, 
the subgraph sampler must have the capability to customize the selection of drug subgraphs for different queries, 
thereby ensuring precise contextualization of the reasoning evidence.
\begin{wraptable}{r}{0.45\textwidth}\small
	\centering
	\small
	\setlength\tabcolsep{3.5pt}
	\caption{Comparing with existing methods."-" represents not applicable.}
	\begin{tabular}{ccc}
		\toprule
		Method & \makecell{Fine-grained \\ Subgraph \\ Selection} & \makecell{Data-specific \\ Encoding  \\ Function} \\
		\midrule
		SEAL~\cite{zhang2018link} & \XSolidBrush & \XSolidBrush \\                     
		GraIL~\cite{teru2020inductive} & \XSolidBrush & \XSolidBrush  \\
		SumGNN~\cite{yu2021sumgnn} & \XSolidBrush & \XSolidBrush \\
		SNRI~\cite{xu2022subgraph} & \XSolidBrush & \XSolidBrush \\
		KnowDDI~\cite{wang2024accurate} & \Checkmark & - \\
		\midrule
		MR-GNAS~\cite{zheng2022multi} & - & \Checkmark \\
		AutoGEL~\cite{wang2021autogel} & - & \Checkmark  \\
		\midrule
		\textbf{CSSE-DDI} & \Checkmark & \Checkmark \\
		\bottomrule
	\end{tabular}
	\label{tab-comparison}
	\vspace{-15px}
\end{wraptable}
Without customized subgraph selection, 
SumGNN samples subgraphs using a fixed subgraph range $k$,
selecting the $k$-hop neighbors of each drug as associated subgraphs for predicting drug-drug interactions (DDIs).
This coarse-grained approach is straightforward and easy to implement,
but it may introduce noise or omit valuable information needed to reason about diverse drug pair interactions.

In terms of encoding process,
the encoding function must be capable of modeling a wide variety of drug interactions within the drug interaction network. 
Real-world drug interactions exhibit complex mechanisms, 
for instance, 
metabolism-based interactions display asymmetric semantic patterns, 
whereas phenotype-based interactions are symmetric.
Manually designed encoding functions are limited in their ability to accommodate both types of distinct semantic patterns simultaneously~\cite{zhang2022bilinear}. 
Therefore, 
designing a customized and data-adaptive subgraph-based pipeline is essential for effective DDI prediction.

Neural architecture search (NAS)~\cite{shen2018automated,elsken2019neural} has achieved remarkable success in designing data-specific models, 
often surpassing architectures crafted by human experts in various fields,
such as computer vision~\cite{zhang2022searching},
graph neural network~\cite{wei2023neural}, 
and knowledge graph learning~\cite{zhang2022bilinear}.
However, 
effectively selecting suitable subgraphs from the vast space of candidates and efficiently optimizing the joint search process of subgraph selection and encoding remain open challenges.

In this paper, 
we leverage NAS to search for data-specific components in the subgraph-based pipeline.
Specifically,
we design search spaces for pipeline components, 
including subgraph selection and encoding spaces,
to capture various drug interaction patterns. 
To enable efficient exploration of the extensive subgraph selection space, 
we introduce a relaxation mechanism that continuously selects subgraphs in a structured manner.
Additionally, 
we propose a subgraph representation approximation strategy to reduce the high cost of explicit subgraph sampling,
enabling efficient and robust search. 
Compared with existing methods in Table~\ref{tab-comparison},
our proposed \textbf{C}ustomized \textbf{S}ubgraph \textbf{S}election and \textbf{E}ncoding for \textbf{D}rug-\textbf{D}rug \textbf{I}nteraction prediction (\textbf{CSSE-DDI}) achieves fine-grained subgrpah selection and data-specific encoding functions,
providing an efficient and precise method for drug interaction prediction.
Our main contributions are summarized as follows:
\begin{itemize}[leftmargin=*]
	\item We present CSSE-DDI,
	a searchable framework for DDI prediction that adaptively customizes the subgraph selection and encoding processes.
	To the best of our knowledge, 
	 this is the first application of NAS techniques to tailor an adaptive subgraph-based pipeline for the DDI prediction task. 
	
	\item We construct expressive search spaces to ensure precise capture of evidence for drug interaction prediction.
	Additionally, 
	we devise a relaxation mechanism to transform the discrete subgraph selection space into a continuous form, 
	enabling differentiable search.
	Simultaneously, 
	we apply a subgraph representation approximation strategy to mitigate the inefficiencies of explicit subgraph sampling,
	thereby accelerating the search process.
	
	\item Extensive experiments on benchmark datasets demonstrate that our method, 
	which searches for customized pipelines, 
	achieves superior performance compared to hand-designed methods. 
	Additionally, our approach effectively captures the underlying biological mechanisms of drug-drug interactions.
\end{itemize}

\section{Related Works}
\vspace{-8pt}
\paragraph{Subgraph-based Link Prediction}
Recently, 
subgraph-based methods~\cite{zhang2018link,teru2020inductive,mai2021communicative} have emerged as a promising approach, 
showing superior performance in link prediction tasks.
Different from canonical GNNs, 
subgraph-based method extracts a subgraph patch for each training and test query,
learning a representation of the extracted patch for final prediction, 
as illustrated in Figure~\ref{fig-pipeline}. 

Existing works has primarily focused on designing more informative subgraphs and more expressive encoding functions.
However, 
they do not take into account customizing these components to deal with various data.
Specifically,
in terms of subgraph sampler, 
current approaches lack fine-grained and adaptive extraction for different query subgraphs.
While PS2~\cite{tan2023bring} demonstrates the effectiveness of identifying optimal subgraphs for each edge in homogeneous graph link prediction, 
there is no comparable work in multi-relational graph link prediction. 
In dense DDI networks, 
fine-grained identification of subgraph for different queries is even more crucial.

As for the encoding function,
existing works overlook the importance of data-specific encoding,
which has been emphasized in recent literature~\cite{wang2022search,wei2023neural}.
Customized encoding functions are especially advantageous for drug interaction networks with complex and diverse interactions.
\vspace{-8pt}
\paragraph{GNN-based DDI Prediction}
Recently, 
there has been growing interest in applying GNNs for DDI prediction~\cite{zitnik2018modeling,huang2020skipgnn}. 
However, 
these works execute message-passing functions over the entire graph, 
which limits their ability to capture explicit local evidence for specific query drug pairs and lack interpretability.
In contrast,
subgraph-based DDI prediction methods~\cite{yu2021sumgnn,hong2022lagat,zhang2023emerging,wang2024accurate} transform the multi-relational link prediction problem into a subgraph classification problem by extracting subgraphs around query nodes, 
achieving strong performance.
Nevertheless,
these works use the same subgraph extraction strategy for all queries and rely on a fixed message-passing function to handle complex DDIs, 
which limits their flexibility and adaptivity in dense DDI networks. 

\vspace{-8pt}
\paragraph{Graph Neural Architecture Search}

Graph neural architecture search (GNAS)~\cite{zhang2021automated} aims to find high-performing GNN architectures using NAS techniques.
Recent studies~\cite{wang2022search,wei2023neural} have explored GNAS to create more expressive GNN models across various tasks.
AutoDDI~\cite{gao2024autoddi}, 
for instance, 
automatically designs GNN architectures to learn molecular graph representations of drugs for DDI prediction. 
However, research on optimizing graph sampling for GNAS remains limited due to the diversity of graph-structured data.

Regarding search strategy, 
early approaches explores the search space using reinforcement learning~\cite{lai2020policy} or evolutionary algorithms~\cite{li2020autograph}, 
which is highly inefficient.
One-shot approaches~\cite{huan2021search} instead construct an over-parameterized network (supernet) and optimize it using gradient descent,
leveraging continuous relaxation of the search space to improve search efficiency.
The recently proposed few-shot NAS paradigm~\cite{zhao2021few} further enhances supernet evaluation consistency by generating multiple sub-supernets. 

\vspace{-8pt}
\section{Proposed Method}
\vspace{-8pt}

\subsection{Problem Formulation}\label{sec-problem-formulation}

Given a set of drugs $\mathcal{V}$ and interaction relations $\mathcal{R}$ among them,
the drug interaction network is denoted as $\mathcal{G}_\text{DDI}=\{(u,r,v)~|~u,v\in\mathcal{V},r\in \mathcal{R}\}$, 
with each tuple $(u,r,v)$ describes an interaction between drug $u$ and drug $v$.
Consequently,
drug-drug interaction (DDI) prediction can be framed a multi-relational link prediction task within the drug interaction network $\mathcal{G}_\text{DDI}$. 
The objective is to predict the types of interactions between two given drug nodes, 
which can be denoted as a query $(u, ?, v)$,
i.e., given the query drug-pair entities $u$ and $v$, 
to determine the interaction $r$ that makes $(u, r, v)$ valid.

Moreover,
instead of directly predicting on the entire graph $\mathcal{G}_\text{DDI}$,
subgraph-based methods decouple the prediction process into two stages:
(1) selecting a query-specific subgraph and (2) encoding the subgraph to predict interactions, 
as shown in Figure~\ref{fig-pipeline}. 
The prediction pipeline then becomes
\begin{equation}	
	\mathcal{G}_\text{DDI} \xmapsto{\texttt{Selection}, (u, v)} \mathcal{G}_{u,v} \xmapsto{\texttt{Encoding}} \boldsymbol{Y}_{u,v},
\end{equation}
where the sampler selects a subgraph $G_{u,v}$ conditioned
on the given query $(u, ?, v)$. 
Using this subgraph $G_{u,v}$, 
the encoding function produces the final predictions $\boldsymbol{Y}_{u,v}$.

Building on previous analysis and existing research, 
and inspired by NAS,
we propose to search for data-adaptive subgraph selection and encoding components to obtain a customized subgraph pipeline.
In Section~\ref{sec-search-space},
we first introduce the well-designed subgraph selection and encoding spaces to ensure comprehensive coverage of cricual information in various drug interaction networks.
Further, 
in Section~\ref{sec-search-strategy}
we present a subgraph relaxation strategy and approximation mechanisms for subgraph representations to facilitate efficient differentiable search.
Finally, 
we develop a robust search algorithm to address the customized search problem with stability and precision.
\vspace{-8pt}
\subsection{Search Space}\label{sec-search-space}
\vspace{-8pt}
\subsubsection{Subgraph Selection Space}

In practice, 
subgraph-based methods define the drug-pair subgraph between drug pairs as the union or interaction of $k$-hop ego-network
\footnote{A $k$-hop ego-network of a node consists of the node and its $k$-hop neighbors.}
of query drugs.
Here, 
$k$ is a key hyperparameter that determines the range of message propagation aggregated by the central node.
Selecting $k$ is crucial to model performance,
as it dictates whether the model has access to high-quality evidence context for accurate prediction.

Prior works~\cite{yu2021sumgnn,wang2023accurate} typically utilize a fixed hyperparameter for all drug pairs, 
i.e., 
selecting the union of a fixed $k$-hop ego-network for arbitrary queries.
Nevertheless, 
this approach can lead to an imprecise collection of evidence for interaction reasoning,
potentially undermining the reasoning process due to missing critical information or the inclusion of excessive irrelevant information.

Based on the above analysis,
we define a drug-pair subgraph selction space containing a range of subgraphs of different sizes for a given query $(u,v)$:
\begin{equation}
	\mathcal{S}_{u,v}=\{\mathcal{G}_{u,v}^{i,j}~|1\leq i,j\leq \eta\},
\end{equation}
where $\mathcal{G}_{u,v}^{i,j}$ is generated by taking the union of the $i$-hop ego-network of node $u$ and the $j$-hop ego-network of node $v$, 
i.e.,
$\mathcal{G}_{u,v}^{i,j}=\{z \in \mathcal{V}~|~z \in (u \cup \mathcal{N}_i(u) \cup v \cup \mathcal{N}_j(v))\}$, 
where $\mathcal{N}_i(u)$ and $\mathcal{N}_j(v)$ are the $i$-hop and the $j$-hop neighbors of $u$ and $v$, respectively.
The threshold $\eta$ constrains the maximum subgraph range.

Since each drug-pair has a specific subgraph selection space,
the overall size of space in the entire graph is $\eta^{2|\mathcal{E}|}$, 
where $|\mathcal{E}|$ represents the number of edges in the drug interaction network.
A larger $|\mathcal{E}|$ result in a subgraph selection space that grows exponentially with the number of edges.
Therefore,
efficiently searching for the optimal subgraph configurations for different queries is challenging.

\subsubsection{Subgraph Encoding Space}

For the automated design of the subgraph encoding function, 
we first adopt a unified message passing framework~\cite{zheng2022multi,di2023message} comprising several key modules: 
the message-computing function $\texttt{MES}$, 
the aggregation function $\texttt{AGG}$, 
the combination function $\texttt{COM}$,
and the activation function $\texttt{ACT}$, 
as follows:
\begin{equation}
	\label{eq:GNN}
	\begin{aligned}
		\texttt{step 1:}& ~~\mathbf{m}_u \gets \texttt{AGG}({\texttt{MES}(\mathbf{h}_v,\mathbf{h}_{r(u,v)})}_{v\in\mathcal{N}_1(u)}),\\ 
		\texttt{step 2:}& ~~\mathbf{h}_u \gets \texttt{ACT}(\texttt{COM}(\mathbf{h}_u,\mathbf{m}_u)),
	\end{aligned}
\end{equation}
where $\mathbf{h}_u\in\mathbb{R}^{d}$ and $\mathbf{h}_{r}\in\mathbb{R}^{d}$ represent the embeddings of node $u$ and interaction $r$, respectively,
and $\mathbf{m}_{u}$ is the intermediate message representation of $u$ aggregated from its neighborhood $\mathcal{N}_1(u)$.

A substantial amount of literature~\cite{xu2018powerful,you2020design,corso2020principal} has focused on manually designing these modules to improve performance.
However,
such encoding functions are inflexible for handling diverse interaction patterns across different drug interaction network.
For example,
interactions in DrugBank~\cite{wishart2018drugbank} describe how one drug affects the metabolism of another one.
The excretion of Acamprosate,
for instance, 
may be decreased when combined with Acetylsalicylic acid (Aspirin).
Such interaction pattern is asymmetric, 
meaning $r(x,y)\nRightarrow r(y,x)$.
Conversely,
interactions in the TWOSIDES dataset~\cite{tatonetti2012data} are primarily at the phenotypic level,
such as headache or pain in throat,
representing symmetric patterns where $r(x,y)\Rightarrow r(y,x)$.
These two relational semantics are distinctly different, 
and existing hand-designed encoding functions struggle to capture such diverse semantics effectively~\cite{shimin2021efficient,zhang2022bilinear}.

Here, 
we aim to perform an adaptive searching for the encoding function in the context of drug interaction prediction.
Based on the framework presented in Eq.~(\ref{eq:GNN}), 
we design an expressive subgraph encoding space with a set of candidate operations.
Detailed explanations of these modules can be found in the Appendix~\ref{apd-search-space}.

After encoding the subgraph $\mathcal{G}_{u,v}$, 
we obtain the representation $\mathbf{z}_{u,v}$ of the input subgraph~$\mathcal{G}_{u,v}$. 
The predictor then maps the representation $\mathbf{z}_{u,v}$ to the probability logits for different interactions between drug pairs, 
i.e., $y_{u,v}=\mathbf{W}_\text{pred}\mathbf{z}_{u,v}$,
where $\mathbf{W}_\text{pred}\in\mathbb{R}^{2d\times |\mathcal{R}|}$ is the parameter of the predictor.

\vspace{-8pt}
\subsection{Search Strategy}\label{sec-search-strategy}
\vspace{-8pt}
\subsubsection{Search Problem}\label{sec-search-problem}
Based on the well-designed search space described above, 
we formulate a bi-level optimization problem to adaptively search for the optimal configuration of subgraph-based pipelines.

\begin{definition}[Customized Subgraph-based Pipeline Search Problem]\label{def-pro}
	Let $\mathcal{A}$ denote the subgraph encoding space,
	$\mathcal{S}_{u,v}$ represent the subgraph selection space for the query $(u,v)$,
	$\bal$ be a candidate encoding function in $\mathcal{A}$, 
	$\mathbf{W}$ represent the parameters of a model from the search space,
	and $\mathbf{W}^*(\mathcal{G}_{u,v};\bal)$ denote the trained operation parameters. 
	Let $\mathcal{D}_\mathrm{tra}$ and $\mathcal{D}_\mathrm{val}$ denote the training and validation sets, respectively.
	The search problem is formulated as follows:
	\begin{align}
		&\argmax\nolimits_{\substack{\bal\in\mathcal{A},  \mathcal{G}_{u,v}\in\mathcal{S}_{u,v}}}\sum\nolimits_{(u,r,v)\in\mathcal{D}_\mathrm{val}}\mathcal{M}(\mathbf{W}^*(\mathcal{G}_{u,v};\bal);\mathcal{G}_{u,v};\bal), \label{upper}\\
		&\text{\;s.t.\;} \mathbf{W}^*(\mathcal{G}_{u,v};\bal)=\!\argmin\nolimits_{\mathbf{W}}\sum\nolimits_{ (u,r,v)\in\mathcal{D}_\mathrm{tra}}\!\!\!\mathcal{L}(\mathbf{W};\mathcal{G}_{u,v};\bal),\label{lower}
	\end{align}
	where the classification loss $\mathcal{L}$ is minimized for all interactions, 
	while the performance measurement $\mathcal{M}$ is expected to be maximized.
\end{definition}
\vspace{-6pt}
In this work, 
we adopt the differentiable search paradigm~\cite{liu2019darts} to solve the bi-level optimization problem, 
which is widely used in recent NAS literature~\cite{dong2024automated} and enables efficient exploration of the search space.
Nevertheless,
our proposed subgraph selection space poses two technical challenges:
\textbf{First}, 
we cannot directly apply relaxation strategies, 
which is a prerequisite for differentiable NAS methods,
to make the discrete selection space continuous.
This limitation arises because different subgraphs in the selection space contain diverse nodes and edges, 
making it challenging to design a relaxation function that unifies subgraphs of varying sizes.
\textbf{Second},
to enable searching within the subgraph selection space, 
we would need to first generate all subgraphs in the space. 
However,
sampling such a large number of subgraphs is computationally intractable.

To address these challenges,
we design a subgraph selection space relaxation mechanism in Section~\ref{sec-ssr} .
Additionally,
we introduce an intuitive subgraph representation approximation strategy in Section~\ref{sec-sas} to reduce the high costs associated with explicit sampling.

\subsubsection{Relaxation of Subgraph Selection Space}\label{sec-ssr}

Technically, 
as in existing NAS works~\cite{liu2019darts,xie2018snas}, 
one typically needs to relax the search space into continuous form to enable effective backpropagation training. 
However, 
for the subgraph selection space, 
the traditional continuous relaxation strategy is not directly applicable due to the structural mismatch between graphs and vectors.

To address this,
we first utilize encoding function $f(\cdot)$ to encode subgraphs with different scopes.
This approach provides all subgraphs with representations of the same dimension, 
making it feasible to implement a relaxation strategy. 
Additionally, 
inspired by the reparameterization trick~\cite{jang2017categorical}, 
we adopt the Gumbel-Softmax function to facilitate differentiable learning over a discrete space:
\begin{equation}
	\hat{\mathbf{z}}_{u,v}^{i,j} = \sum_{1\leq i,j\leq\eta}\frac{\exp(\log (g(f(\mathcal{G}_{u,v}^{i,j})) + \textbf{G}_{i,j})/\tau)}{\sum_{i',j'=1}^\eta \exp(\log (g(f(\mathcal{G}_{u,v}^{i',j'})) + \textbf{G}_{i',j'})/\tau)}f(\mathcal{G}_{u,v}^{i,j}),
\end{equation}
where $g(\cdot)$ scores the subgraph representations using multiple linear layers,
$\textbf{G}_{i,j}=-\log(-\log(\textbf{U}_{i,j}))$ is the Gumbel random variable, 
$\textbf{U}_{i,j}$ is a uniform random variable, 
and $\tau$ is the temperature parameter controlling sharpness.
$\hat{\mathbf{z}}_{u,v}^{i,j}$ is the mixed selection operation of subgraph $\mathcal{G}_{u,v}^{i,j}$ used to optimize searching process.

\subsubsection{Subgraph Representation Approximation Strategy}\label{sec-sas}

To solving the optimization problem as Eq.~(\ref{upper}) and (\ref{lower}),
we need to explicitly sample all the candidate subgraphs within the subgraph selection space $\mathcal{S}_{u,v}$ for each query. 
However,
one of the most challenging aspects of subgraph-based approaches is their inefficient subgraph sampling process~\cite{zhang2023adaprop,zhang2022knowledge,zhou2024less}.

Upon examining our subgraph selection space, 
we observe that all subgraphs are generated by combining multi-hop ego-networks of the target nodes, 
encompassing multiple neighborhood hops.
Inspired by the $k$-subtree extractor~\cite{chen2022structure}, 
we apply an encoding function to the entire graph and use the resulting node representations of $u$ and $v$ as the ego-network representations for these nodes. 
The representation of the drug pair can then be obtained by concatenating the ego-network representations of $u$ and $v$.
Formally, 
if we denote by $f(\mathcal{G}_\text{DDI},u,i)$ the $i$-layer hidden representation of node $u$ produced by encoding function applied to $\mathcal{G}_\text{DDI}$, then
\begin{equation}
	f(\mathcal{G}_{u,v}^{i,j}) \approx \texttt{CONCAT}(f(\mathcal{G}_\text{DDI},u,i),f(\mathcal{G}_\text{DDI},v,j)),
\end{equation}
The $k$-subtree extractor represents the $k$-subtree structure rooted at a given node, 
which mirrors the structure as the $k$-hop ego-network.
This approximation strategy only requires executing the encoding function on the entire drug interaction network, 
thereby efficiently yielding subgraph representations of varying scopes, 
which significantly improves the efficiency in solving the bi-level optimization problem.

\vspace{-8pt}
\subsubsection{Robust Search Algorithm}

Using the proposed subgraph selection relaxation mechanism, 
we can transform the overall discrete search space in Definition.~\ref{def-pro} into a continuous form,
allowing the search problem to be solved by the one-shot NAS paradigm.
Additionally, 
our subgraph representation approximation strategy efficiently obtains subgraph representations and reduces search costs

Following~\cite{zhang2023autogt},
we adopt the single path one-shot training strategy (SPOS)~\cite{guo2020single} to reduce the computational cost of supernet training.
However, 
the one-shot approach~\cite{brock2018smash,pham2018efficient,liu2019darts}, 
i.e., using the same supernet parameters $\bW$ for all architectures, 
can decrease the consistency between the supernet's performance estimation and the ground-truth performance~\cite{yu2020evaluating}. 
Inspired by few-shot NAS~\cite{zhao2021few}, 
we propose a message-aware partitioned supernet training strategy to mitigate the coupling effect of different message-computing operators~\cite{tan2023kracl}.
By partitioning the superent to form sub-supernets based on the type of message-computing function, 
this strategy improves the consistency and accuracy of supernet,
enabling the search algorithm more stable and robust.
Algorithm~\ref{algo:csseddi_main} delineates the full procedure,
with further details provided in Appendix~\ref{apd-search-algorithm}.
\vspace{-8pt}

\begin{algorithm}[ht]
	\SetAlgoLined
	\small
	\KwIn{Supernet $\mathcal{S}$, number of partitions based on message computing function categories $M$ $(M=4)$, subsupernet $\mathcal{S}_i, (i=1,\cdots,M)$.}
	\tcp*[h]{supernet training phase}\\
	Train $\mathcal{S}$ by continuously sampling a single path until convergence;\\
	\tcp*[h]{supernet partition phase}\\
	Partition $\mathcal{S}$ into $M$ sub-supernets $\mathcal{S}_1, \cdots, \mathcal{S}_M$; \\
	\tcp*[h]{sub-supernet training phase}\\
	\ForAll{$i = 1, \cdots, M$ }{
		Initialize $\mathcal{S}_i$ with weights transferred from $\mathcal{S}$; \\
		Train $\mathcal{S}_i$ by continuously sampling a single path until convergence;\\
	}
	\tcp*[h]{searching phase}\\
	Search the optimal encoding function from sub-supernets $\mathcal{S}_1, \cdots, \mathcal{S}_M$ on validation data by natural gradient descent;\\
	Select the optimal subgraphs from sub-supernets $\mathcal{S}_1, \cdots, \mathcal{S}_M$ on validation data by preserving the subgraphs with the largest probabilities;
	\caption{The search algorithm of CSSE-DDI.}
	\label{algo:csseddi_main}
\end{algorithm}
\vspace{-15pt}
\subsection{Comparison with Existing Works}

While many works~\cite{yu2021sumgnn,zhang2023emerging,wang2024accurate} have explored DDI prediction using subgraph-based methods,
our approach introduces two significant advancements.
First,
to the best of our knowledge, 
our method (CSSE-DDI) is the first to customize the subgraph selection and encoding processes specifically for subgraph-based DDI prediction.
In contrast, 
previous methods rely on fixed subgraph selection strategy to sample subgraphs and employ hand-designed functions for encoding,
as summarized in Table~\ref{tab-comparison}.
Consequently,
our method can adapt data-specific components within subgraph-based pipelines,
outperforming existing methods in both performance and efficiency (Section~\ref{sec-perf}).
Moreover,
our approach not only selects fine-grained drug-pair subgraphs that enhance interpretability through potential pharmacokinetic and metabolic concepts (Section~\ref{sec-ss}), 
but also searches for data-specific encoding functions that accurately capture the semantic features of drug interactions (Section~\ref{sec-sf}).

\vspace{-8pt}
\section{Experiments}
\vspace{-8pt}
\subsection{Experimental Setup}
\paragraph{Datasets}
Experiments are conducted on two public benchmark DDI datasets: DrugBank~\cite{wishart2018drugbank} and TWOSIDES~\cite{tatonetti2012data}. 
Detailed descriptions of these datasets are presented in Appendix~\ref{apd-dataset}.
\vspace{-8pt}
\paragraph{Experimental Settings}
Following~\cite{zhang2023emerging}, 
we examine two DDI prediction task settings: S0 and S1.
Let the drug pairs for DDI prediction be denoted as $(u, v)$.
In the S0 setting, 
both drug nodes $u$ and $v$ are present in the known DDI graph.
Existing DDI prediction methods are typically evaluated in this setting.
In contrast, 
the S1 setting involves a pair (u, v) where one drug is known and the other is a novel drug not represented in the known DDI graph. 
This scenario highlights the critical need for DDI predictions involving new drugs in real-world applications.
\vspace{-8pt}
\paragraph{Evaluation Metric}
We follow~\cite{yu2021sumgnn} to evaluate our method. 
For the DrugBank dataset,
where each drug pair contains only one interaction,  
we use the following metrics: F1 Score, Accuracy and Cohen’s $\kappa$.
For the TWOSIDES dataset, 
where multiple interactions may exist between a pair of drugs, 
we consider the following metrics: ROC-AUC, PR-AUC and AP@50.
Additional details are provided in Appendix~\ref{apd-eva}.
\vspace{-8pt}
\paragraph{Baselines} 
We compare CSSE-DDI with the following representative DDI prediction method:
(i) GNN-based methods include Decagon~\cite{zitnik2018modeling}, GAT~\cite{velivckovic2018graph}, 
SkipGNN~\cite{huang2020skipgnn},
CompGCN~\cite{vashishth2020composition},
ACDGNN~\cite{yu2023attention},
and TransFOL~\cite{cheng2024transfol}.
(ii) Subgraph-based methods include SEAL~\cite{zhang2018link}, 
GraIL~\cite{teru2020inductive}, 
SumGNN~\cite{yu2021sumgnn},
SNRI~\cite{xu2022subgraph},
KnowDDI~\cite{wang2023accurate}
and LaGAT~\cite{hong2022lagat}.
(iii) NAS-based method include
MR-GNAS~\cite{zheng2022multi},
and AutoGEL~\cite{wang2021autogel}.

We also compare our method with two variants, 
including CSSE-DDI-FS and CSSE-DDI-FF.
The configurations of these variants are as follows:
(i) \textbf{CSSE-DDI-FS}: This variant omits fine-grained subgraph selection for each query, using fixed k-layer drug node representations to generate the subgraph representation.
(ii) \textbf{CSSE-DDI-FF}: This variant does not search for the encoding function, instead using a fixed encoding function backbone to capture semantic and topological features in the drug interaction network. In this case, we employ a 3-layer CompGCN model as the backbone.
For all baselines, 
we obtain the results by rerunning the released codes. 
\vspace{-8pt}
\paragraph{Implementation}
We implement our method\footnote{Our code is available at \href{https://github.com/LARS-research/CSSE-DDI}{https://github.com/LARS-research/CSSE-DDI}.} based on PyTorch framework~\cite{paszke2019pytorch}.
Following existing GNN-based methods~\cite{wang2023accurate}, 
we select a 3-layer encoding function backbone for both datasets.
The maximum threshold~$\eta$ for the subgraph selection space is set to 3.
More experimental details are given in the Appendix~\ref{apd-implement}.

\vspace{-8pt}
\subsection{Performance Comparison in S0 settings}\label{sec-perf}

Table~\ref{tab:main} shows the overall results across all benchmarks in S0 setting.
As can be seen, 
CSSE-DDI consistently outperforms all baselines on each dataset, 
demonstrating its effectiveness in searching for data-specific subgraph-based pipelines for DDI prediction task.
Among the baselines, 
subgraph-based methods significantly outperform full-graph-based methods due to their enhanced ability to reason over local subgraph contexts.
Within the subgraph-based methods,
SEAL,
GraIL,
SumGNN,
and SNRI use a fixed sample strategy to select subgraphs,
which may not be optimal for different drug-pair queries.

When it comes to NAS-based method,
MR-GNAS and AutoGEL contain well-established search spaces that embrace multi-relational message-passing schema,
focusing primarily on automated encoding function design using the one-shot NAS paradigm. 
While CSSE-DDI adopts a single path supernet training strategy and a message-aware partitioning approach to search for data-adaptive subgraph-based pipelines with stability and robustness,
enabling the model to achieve excellent performance across various datasets.
Moreover,
the consistent performance gains of CSSE-DDI over its two variants validate the importance of jointly customizing subgraph-based pipeline components, 
i.e., fine-grained subgraphs and data-specific encoding functions,
to fit datasets rather than relying on a fixed approach.

\begin{table}[t]
	\centering
	\setlength\tabcolsep{3.5pt}
	\caption{CSSE-DDI achieves the best predictive performance compared to state-of-the-art baselines in DDI prediction. Average and standard deviation of five runs are reported. For these metrics, higher values always indicate better performance. }
		\begin{tabular}{cccccccc}
			\toprule
			\multirow{3}{*}{\makecell{\textbf{Model}\\ \textbf{Type}}}&\textbf{Dataset} &\multicolumn{3}{c}{\textbf{Dataset 1: DrugBank}} & \multicolumn{3}{c}{\textbf{Dataset 2: TWOSIDES}} \\ \cmidrule(lr){2-8}
			&\textbf{Task Type} &\multicolumn{3}{c}{Multi-class} & \multicolumn{3}{c}{Multi-label} \\ \cmidrule(lr){2-8}
			&\textbf{Methods} & F1 Score & Accuracy & Cohen’s $\kappa$ & ROC-AUC & PR-AUC & AP@50\\ 
			\midrule
			\multirow{6}{*}{\makecell{GNN-\\based}}&Decagon & 57.35\std0.26 & 87.19\std0.28 & 86.07\std0.08 & 91.72\std0.04 & 90.60\std0.12	& 82.06\std0.45 \\
			&GAT     & 33.49\std0.36 & 77.18\std0.15 & 74.20\std0.23 & 91.18\std0.14 & 89.86\std0.05 & 82.80\std0.17 \\
			&SkipGNN & 59.66\std0.26 & 85.83\std0.18 & 84.20\std0.16 & 92.04\std0.08 & 90.90\std0.10	& 84.25\std0.25 \\
			&CompGCN & 71.20\std0.70 & 88.30\std0.29 & 86.15\std0.35 & 93.00\std0.07 & 91.26\std0.07	& 86.18\std0.10 \\
			&ACDGNN & 86.24\std0.93& 90.53\std0.38 &87.81\std0.33 &93.69\std0.47&92.12\std0.21& 87.45\std0.24\\
			&TransFOL& 89.97\std1.64&91.92\std0.89& 90.92\std0.72&94.16\std0.62&93.52\std0.53& 88.13\std0.39\\
			\midrule
			\multirow{6}{*}{\makecell{Subgraph-\\based}}&SEAL   & 48.82\std0.98 & 76.61\std0.26 & 71.91\std0.59 & 90.74\std0.22 & 90.11\std0.17 & 84.13\std0.13 \\
			&GraIL   & 73.20\std0.69 & 85.40\std0.39 & 82.70\std0.47 & 92.93\std0.10 & 91.69\std0.14 & 87.43\std0.09 \\
			&SumGNN  & 78.35\std0.51 & 89.05\std0.36 & 87.28\std0.08 & 92.62\std0.04 & 90.80\std0.40	& 85.75\std0.10 \\
			&SNRI & 85.57\std0.32 & 90.15\std0.21 & 88.94\std0.36 & 93.12\std0.18 & 92.64\std0.12 & 87.53\std0.11 \\
			&KnowDDI & \underline{90.06\std0.27} & \underline{93.15\std0.19} & \underline{91.87\std0.21} & \underline{95.05\std0.06} & \underline{93.75\std0.05}	& \underline{89.24\std0.06} \\
			&LaGAT&81.63\std0.56&86.21\std0.18& 85.38\std0.23&89.78\std0.21&86.33\std0.15& 83.75\std0.36\\
			\midrule
			\multirow{5}{*}{\makecell{NAS-\\based}}&MR-GNAS & 74.24\std0.45 & 88.17\std0.24 & 87.31\std0.11 & 93.85\std0.07 & 91.80\std0.03	& 87.16\std0.05 \\
			&AutoGEL & 76.87\std0.63 & 89.35\std0.59 & 86.14\std0.41 & 94.11\std0.32 & 92.35\std0.29 & 88.13\std0.41 \\
			\cmidrule(lr){2-8}
			&CSSE-DDI-FS  &86.31\std 0.36& 91.08\std 0.21& 89.17\std 0.27 & 94.35\std0.07 & 93.01\std0.06 & 89.08\std 0.04 \\
			&CSSE-DDI-FF & 80.96\std 0.65& 90.27\std 0.23& 88.69\std 0.31 & 94.26\std0.08 & 92.74\std0.06 & 88.91\std 0.09 \\
			&\textbf{CSSE-DDI} &\textbf{92.08\std 0.22}&\textbf{95.56\std 0.15}& \textbf{94.72\std 0.26} &\textbf{95.47\std0.02} &	\textbf{94.21\std0.05} &\textbf{89.76\std 0.05} \\
			\bottomrule
		\end{tabular}
	\label{tab:main}
	\vspace{-17px}
\end{table}

\begin{wrapfigure}{r}{0.6\textwidth}
	\vspace{-10pt}
	\centering
	\subfloat[DrugBank]{
		\includegraphics[width=0.5\linewidth]{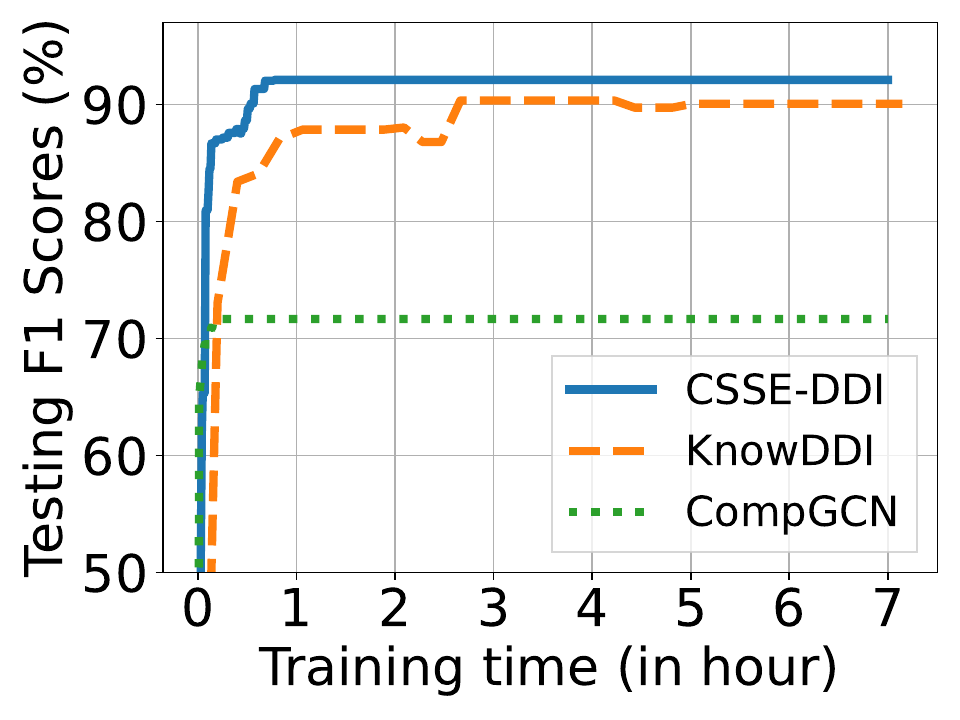}
	}%
	\subfloat[TWOSIDES]{
		\includegraphics[width=0.5\linewidth]{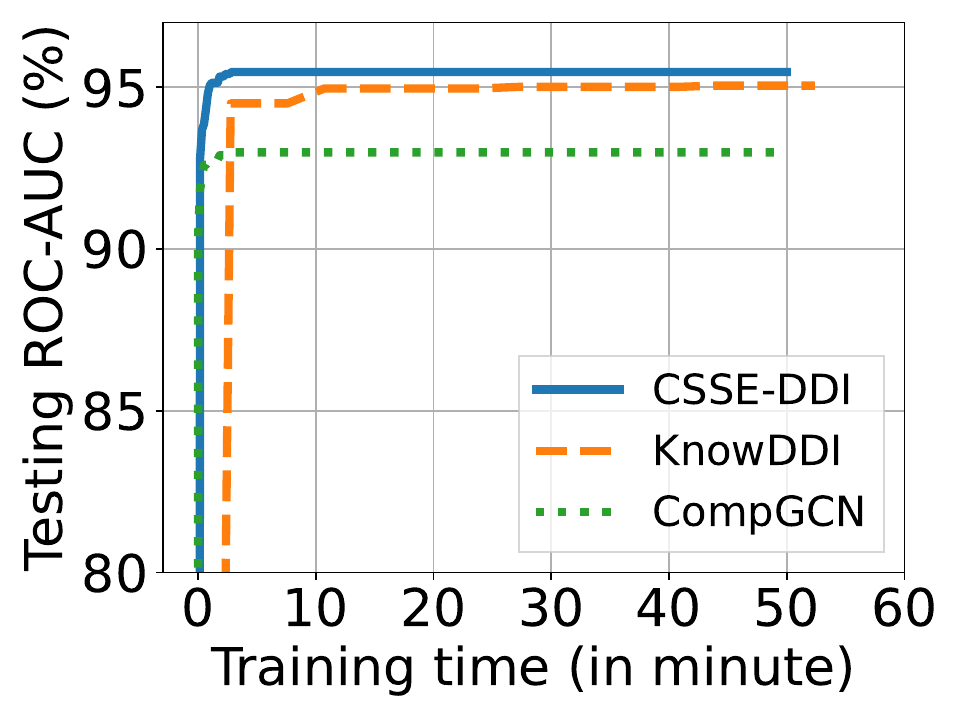}
	}
	\caption{Comparison on convergence between the
		searched architectures by CSSE-DDI and human-designed
		methods.} 
	\label{fig-learning_curve}
	\vspace{-5pt}
\end{wrapfigure}

Figure~\ref{fig-learning_curve} shows the learning curves
of several competitive methods on both datasets,
including CompGCN, KnowDDI and the proposed CSSE-DDI.
As can be seen,
the searched models not only outperform the baselines
but also demonstrate a clear advantage in efficiency, 
highlighting that enhancing model flexibility and adaptability is essential for improving performance and efficiency.

\subsection{Choices of Search Strategy}
\vspace{-5pt}
To demonstrate the effectiveness of our search strategy, 
we introduce two variants with different search strategies:
(i) \textbf{CSSE-DDI w/o MAP}: This variant uses only one trained supernet to serve as a performance evaluator for candidate architectures, instead of generating multiple sub-supernets by Message-Aware Partition (MAP) strategy. 
(ii) \textbf{CSSE-DDI w/o SPOS}: This variant utilizes the message-aware partition strategy to jointly optimize the supernet weights and architectural parameters, without using the Single Path One-Shot (SPOS) strategy~\cite{guo2020single} .

\begin{wraptable}{r}{0.6\textwidth}
	\vspace{-13px}
	\centering
	\setlength\tabcolsep{3.5pt}
	\caption{Performance of CSSE-DDI using different variants of search algorithm.}
	\begin{tabular}{ccc}
		\toprule
		\textbf{Variant} & \textbf{DrugBank} & \textbf{TWOSIDES}          \\ 
		\midrule
		CSSE-DDI w/o MAP & 90.17\std0.29 &  95.12\std0.04   \\
		\midrule
		CSSE-DDI w/o SPOS & 90.97\std0.72 &  94.89\std0.13   \\
		\midrule
		\textbf{CSSE-DDI}  & \textbf{92.08\std0.22} & \textbf{95.47\std0.02} \\
		\bottomrule
	\end{tabular}
	\label{tb-search-algorithm}
	\vspace{-5pt}
\end{wraptable}
In Table \ref{tb-search-algorithm}, 
we compare CSSE-DDI with other variants.
As can be seen,
the absence of either message-aware partition strategy or sampling-based NAS strategy negatively impacts performance.
The performance gains achieved through the message-aware partition strategy arise from using multiple sub-supernets, 
which provide more accurate performance estimations to guide the search process. 
Regarding the SPOS strategy, it decouples supernet training from architecture search, making it more efficient and robust in practice.

\vspace{-8pt}
\subsection{Sensitivity Analysis of the Threshold~$\eta$}
\vspace{-5pt}

\begin{wrapfigure}{r}{0.6\textwidth}
	\vspace{-10px}
	\centering
	\subfloat[DrugBank]{
		\includegraphics[width=0.5\linewidth]{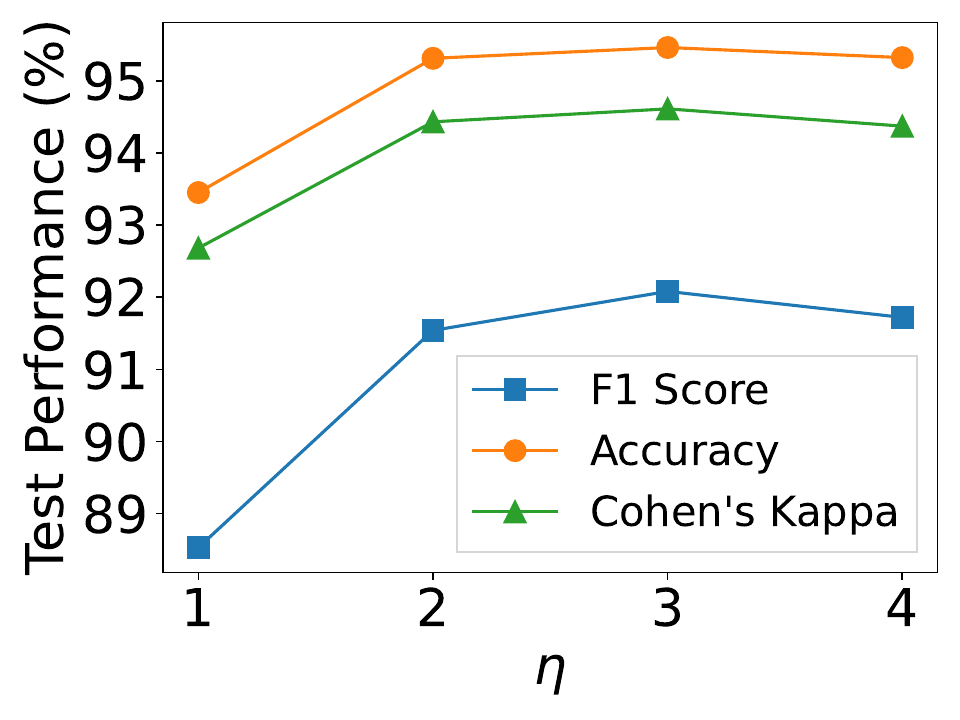}
	}%
	\subfloat[TWOSIDES]{
		\includegraphics[width=0.5\linewidth]{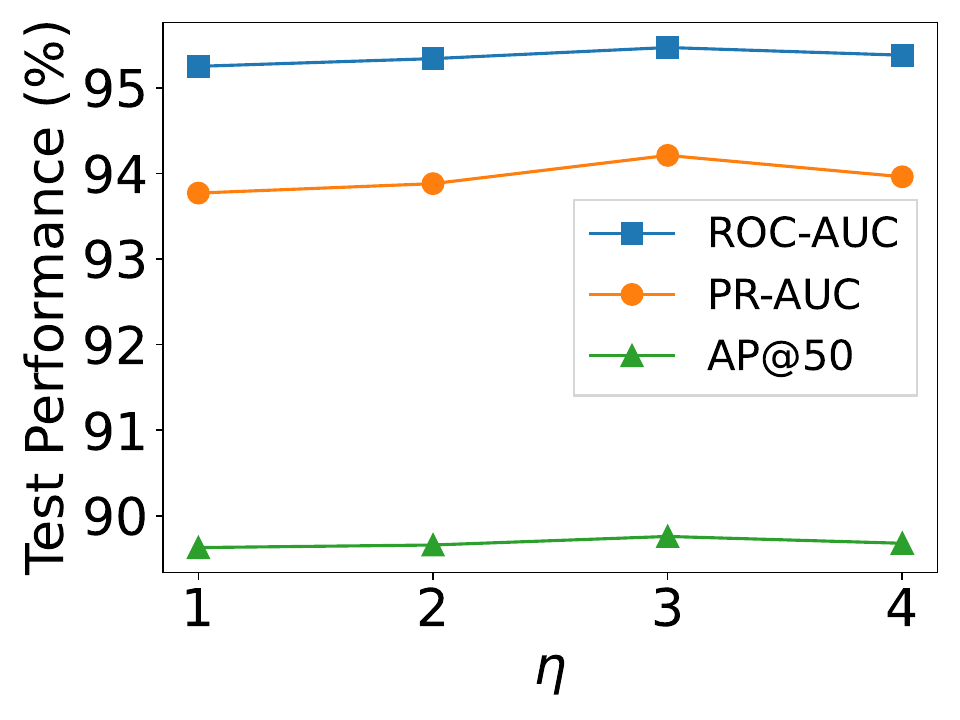}
	}
	\caption{Performance given different hyperparameter~$\eta$.} 
	\label{fig-effect_eta}
	\vspace{-5pt}
\end{wrapfigure}

Here, 
we analyze the effect of the threshold~$\eta$ used in subgraph selection space.
Figure~\ref{fig-effect_eta} shows the impact of varying~$\eta$.
As can be observed,
model performance continues to get better as the threshold~$\eta$ grows. 
When the threshold~$\eta=3$,
the model performance nears saturation,
as larger thresholds do not lead to further improvements. 
This is likely because most of the essential information for DDI prediction is contained within the $3$-hop ego-subgraphs of target drugs.
Intuitively, 
larger subgraphs may provide additional useful information.
However, in practice, due to the inherent biases of the search algorithm, achieving an optimal model may be challenging.
When $\eta$ is too large,
it may introduce noise and dilute the critical information.
A similar phenomenon has been found in the existing work SumGNN~\cite{yu2021sumgnn}.
Besides, excessively large thresholds~$\eta$ will only lead to unnecessary expansion of the search space and higher computational costs.

\subsection{Performance Comparison in S1 settings}
 
To further validate the effectiveness of our method, 
we use the S1 setting in the EmerGNN~\cite{zhang2023emerging} method, 
to predict drug-drug interactions between emerging drugs and existing drugs. 
The experimental results are shown in Table~\ref{tab:s1}. 
A significant performance drop from the transductive setting (S0) to the inductive setting (S1) demonstrates that DDI prediction for new drugs is more challenging.
Although Emergnn, 
which is specifically designed for new drug prediction, 
achieves optimal performance,
CSSE-DDI still demonstrates impressive results, 
outperforming existing GNN-based and subgraph-based methods. 
This strong performance is largely due to the robust learning capability of NAS technology in handling unknown data.
\vspace{-10px}
\begin{table}[h]
	\centering
	\caption{Experimental results in S1 setting.}
	\vspace{5px}
	\resizebox{\textwidth}{!}{
	\begin{tabular}{ccccccc}
		\toprule
		\textbf{Dataset} &\multicolumn{3}{c}{\textbf{Dataset 1: DrugBank}} & \multicolumn{3}{c}{\textbf{Dataset 2: TWOSIDES}} \\ \cmidrule(lr){1-7}
		\textbf{Task Type} &\multicolumn{3}{c}{Multi-class} & \multicolumn{3}{c}{Multi-label} \\ \cmidrule(lr){1-7}
		\textbf{Methods} & F1 Score & Accuracy & Cohen’s $\kappa$ & ROC-AUC & PR-AUC & Accuracy\\ 
		\midrule
		CompGCN & 30.98\std3.26 & 52.76\std0.46 & 37.87\std1.28 & 84.83\std1.02 & 83.68\std1.86	& 74.64\std0.79 \\
		Decagon& 11.39\std0.79 & 32.56\std0.92 & 20.29\std1.33 & 57.49\std1.75 & 59.38\std1.09 & 52.27\std1.48 \\
		SumGNN & 26.57\std1.59 & 44.30\std1.04 & 40.24\std1.26 & 80.02\std2.17 & 78.42\std1.62	& 69.81\std1.77 \\
		KnowDDI & 31.14\std1.24 & 53.44\std1.73 & 43.93\std1.17 & 84.23\std2.63 & 82.58\std1.94	& 74.72\std1.51 \\
		EmerGNN & \textbf{58.13\std1.36}& \textbf{69.53\std1.97} & \textbf{62.19\std1.62} &\underline{87.42\std0.39}&\underline{86.20\std0.71}& \underline{79.23\std0.54}\\
		\midrule
		\textbf{CSSE-DDI}& \underline{37.24\std1.13}&\underline{58.57\std0.85}& \underline{49.97\std1.01}&\textbf{88.33\std0.52}&\textbf{86.47\std0.27}& \textbf{80.01\std0.39}\\
		\bottomrule
	\end{tabular}
	}
	\label{tab:s1}
	\vspace{-17px}
\end{table}

\subsection{Case Study}

\subsubsection{Fine-grained Subgraph Selection}\label{sec-ss}

\begin{wrapfigure}{r}{0.59\textwidth}
	\centering
	\vspace{-15pt}
	\includegraphics[width=1\linewidth]{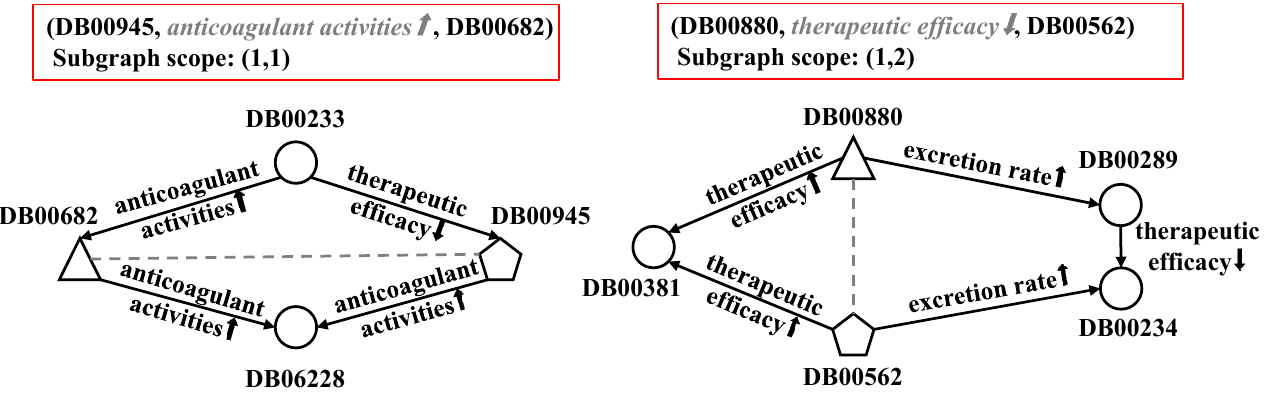}
	\vspace{-5pt}
	\caption{Visualization of the searched subgraphs corresponding to the specific drug pairs.} 
	\label{fig-subgraph}
	\vspace{-10pt}
\end{wrapfigure}
We visualize exemplar query-specific subgraphs from the DrugBank dataset in Figure~\ref{fig-subgraph}, 
highlighting \textbf{domain concepts} such as pharmacokinetics, metabolism, and receptor interactions. 
As shown,
CSSE-DDI can identify distinctive subgraphs containing semantic information to support inference for different queries,
revealing pharmacokinetic and metabolic relationships.

For example, 
to predict the interaction between DB00945 (Aspirin) and DB00682 (Warfarin),
CSSE-DDI searches out the subgraph scope $(1,1)$, 
as depicted on the left part of Figure~\ref{fig-subgraph}.
Firstly, 
it can be seen from the figure that the therapeutic efficacy of DB00233 (Aminosalicylic acid) can decrease when combined with DB00945 (Aspirin),
suggesting similarity between the two drugs~\cite{ariens1970reduction,foucquier2015analysis}
Given that DB00233 (Aminosalicylic acid) may increase the anticoagulant activity of DB00682 (Warfarin) and that DB00233 resembles DB00945 (Aspirin),
it can be inferred that DB00945 (Aspirin) may similarly increase the anticoagulant activity of DB00682 (Warfarin).
This example demonstrates that the identified subgraph contains sufficient semantic information to reason about the interaction between DB00945 (Aspirin) and DB00682 (Warfarin).

\subsubsection{Data-specific Encoding Function}\label{sec-sf}

\begin{wrapfigure}{r}{0.6\textwidth}
	\centering
	\vspace{-10pt}
	\includegraphics[width=1\linewidth]{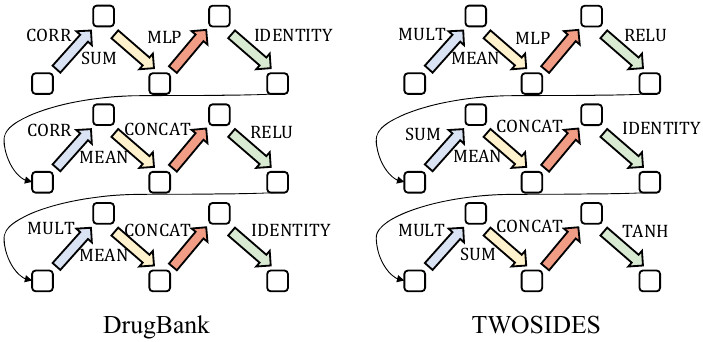}
	\vspace{-5pt}
	\caption{The searched encoding functions on all benchmark datasets.} 
	\label{fig-function}
	\vspace{-10px}
\end{wrapfigure}
Furthermore,
we visualize the searched structure of encoding functions across all datasets in Figure \ref{fig-function}.
It is clearly illustrated that different combinations of the designed operations,
i.e., data-specific encoding functions, 
are obtained.

In particular, 
the searched message-computing functions contain
more \texttt{CORR} operations in the DrugBank dataset, 
while more \texttt{MULT} functions are searched in the TWOSIDES dataset.
The \texttt{CORR} function is non-commutative~\cite{nickel2016holographic}, 
making it suitable for modeling asymmetric interactions (e.g., metabolic-based interactions) present in DrugBank.
While \texttt{MULT} is suitable for modeling symmetric relations (phenotype-based interactions) due to its exchangeability~\cite{yang2015embedding}.

\section{Conclusion}
We propose a searchable framework, CSSE-DDI, for DDI prediction.
Specifically,
we design refined search spaces to enable fine-grained subgraph selection and data-specific encoding function optimization.
To facilitate efficient search, 
we introduce a relaxation mechanism to convert the discrete subgraph selection space into a continuous one.
Additionally, 
we employ a subgraph representation approximation strategy to accelerate the search process, 
addressing the inefficiencies of explicit subgraph sampling.
Extensive experiments demonstrate that CSSE-DDI significantly outperforms state-of-the-art methods. 
Moreover, 
the search results generated by CSSE-DDI offer interpretability in the context of drug interactions, 
revealing domain-specific concepts such as pharmacokinetics and metabolism.

\section*{Acknowledgements}
We thank the anonymous reviewers for their valuable comments.
This work was supported
in part by the National Key Research and Development Program of China (Grant No. 2022ZD0160300), 
in part by the National Science Fund for Distinguished Young Scholars (Grant No. 62025602),
in part by the National Natural Science Foundation of China (Grant Nos. U22B2036, 11931015, 62203363, and 92270106), 
in part by the Technology Innovation Leading Program of Shaanxi (Grant No. 2023GXLH-086),
in part by the Beijing Natural Science Foundation (Grant No. 4242039).
in part by the Fok Ying-Tong Education Foundation, China (Grant No. 171105),
in part by the Fundamental Research Funds for the Central Universities (Grant Nos. G2024WD0151 and D5000240309),
and in part by the Tencent Foundation and XPLORER PRIZE.

\clearpage

\bibliographystyle{unsrtnat}
\bibliography{neurips_2024} 


\onecolumn
\appendix

\part{Appendix} 
\vspace{-20pt}
\etocdepthtag.toc{mtappendix}
\etocsettagdepth{mtchapter}{none}
\etocsettagdepth{mtappendix}{subsection}
\renewcommand{\contentsname}{}
\tableofcontents
\newpage
\section{More Method Details}
\subsection{Subgraph Encoding Space}\label{apd-search-space}
An expressive subgraph encoding space can be naturally designed by including human-designed operations, 
the details of which are given in Table~\ref{tab-search-space}.

\begin{table}[h]
	\centering
	\caption{The operations used in our search space.}
	\begin{tabular}{cc}
		\toprule
		\textbf{Function name} & \textbf{Operations}               \\
		\midrule
		\makecell[c]{Message Computing Function}  & \makecell[c]{\texttt{SUB}, \texttt{MULT},  \\ \texttt{CORR}, \texttt{ROTATE}}                              \\\midrule
		\makecell[c]{Aggregation Function} & \makecell[c]{\texttt{SUM}, \texttt{MAX},\texttt{MEAN}}                                \\\midrule
		\makecell[c]{Combination Function}     & \makecell[c]{\texttt{MLP}, \texttt{CONCAT}}                    \\\midrule
		\makecell[c]{Activation Function}         & \makecell[c]{\texttt{RELU}, \texttt{TANH},\\ \texttt{IDENTITY}} \\
		\bottomrule
	\end{tabular}
	\label{tab-search-space}
\end{table}

In particular,
given the embedding $\mathbf{h}_u$ of node $u$ and the embedding $\mathbf{h}_r$ of interaction $r$, 
the message computing function takes the following form:
$\texttt{MES}_\texttt{SUB}=\mathbf{h}_u-\mathbf{h}_r$,
$\texttt{MES}_\texttt{MULT}=\mathbf{h}_u*\mathbf{h}_r$,
$\texttt{MES}_\texttt{CORR}=\mathbf{h}_u\star\mathbf{h}_r$,
$\texttt{MES}_\texttt{ROTATE}=\mathbf{h}_u\circ\mathbf{h}_r$,
where $\star$ stands for the circular correlation operation~\cite{nickel2016holographic},
$\circ$ represents the rotation operation~\cite{sun2019rotate}.

\subsection{Robust Search Algorithm}\label{apd-search-algorithm}

We adopt the single path one-shot (SPOS) training strategy to solve the customized search problem, 
which decouple supernet training and architecture searching.
In particular, 
definition~\ref{def-pro} can be transformed into a two-step optimaztion~\cite{guo2020single}:
\begin{align}
	&\argmax\nolimits_{\substack{\bal\in\mathcal{A},  \mathcal{G}_{u,v}\in\mathcal{S}_{u,v}}}\sum\nolimits_{(u,r,v)\in\mathcal{D}_\mathrm{val}}\mathcal{M}(\mathbf{W}^*;\mathcal{G}_{u,v};\bal), \label{apx-upper}\\
	&\mathbf{W}^*=\!\argmin\nolimits_{\mathbf{W}}\mathbb{E}_{\bal\in\mathcal{A}}\sum\nolimits_{ (u,r,v)\in\mathcal{D}_\mathrm{tra}}\!\!\!\mathcal{L}(\mathbf{W};\mathcal{G}_{u,v};\bal),\label{apx-lower}
\end{align}
where $\bW$ denotes the shared learnable weights in the supernet with its optimal value $\bW^*$ for all the architectures in the overall search space.

Eq.~(\ref{apx-lower}),(\ref{apx-upper}) represent the supernet training and architecture searching phase, respectively.
In the following, we will describe the detailed process of the two phases.

\subsubsection{Supernet Training}

In supernet training phase,
a sub-model $\bal$ is sampled according to the discrete distribution $\pi(\mathcal{A})$.
Thus, Eq.~(\ref{apx-lower}) can be formulated as
\begin{equation}
	\mathbf{W}^*=\!\argmin\nolimits_{\mathbf{W}}\mathbb{E}_{\bal\sim\pi(\mathcal{A})}\sum\nolimits_{ (u,r,v)\in\mathcal{D}_\mathrm{tra}}\!\!\!\mathcal{L}(\mathbf{W};\mathcal{G}_{u,v};\bal),\label{apx-sp}
\end{equation}
where the discrete distribution $\pi(\mathcal{A})$ is set to uniform distribution.

First, 
we need to perform single path sampling to train the supernet until it converges.
In the next step, 
we need to partition the supernet into sub-supernets.
which is a key step aiming to isolate operations that are coupled with each other.
This allows the supernet to be trained and converge more stably.

In our supernet,
we use a message-aware partitions strategy due to the fact that
the degree of dissimilarity between the operations in the message computing function $\texttt{MES}$ is much higher compared with others. 
These operations focus on capture different semantic types of interactions,
which has been discussed in existing works~\cite{trouillon2016complex,sun2019rotate,tan2023kracl}.
Therefore, 
we partition four operations of the message computing function of the first layer of the supernet, 
to improve the accuracy of the performance estimation.

After partitioning operation,
we initialize four sub-supernets with weights transferred from the original supernet. 
Next, we train these sub-supernets to convergence by sampling single path.
Here,
the supernet training phase is all done.

\subsubsection{Architecture Searching}

After completing sub-supernet training phase,
we have obtained well-trained supernet weights.
In the searching phase, 
Eq.~(\ref{apx-upper}) can be transformed as
\begin{align}
	&\argmax\nolimits_{\mathcal{G}_{u,v}\in\mathcal{S}_{u,v}}\sum\nolimits_{(u,r,v)\in\mathcal{D}_\mathrm{val}}\mathcal{M}(\mathbf{W}^*;\mathcal{G}_{u,v};\bal), \label{apx-ss} \\
	&\text{\;s.t.\;}\argmax\nolimits_{\bal\in\mathcal{A}}\sum\nolimits_{(u,r,v)\in\mathcal{D}_\mathrm{val}}\mathcal{M}(\mathbf{W}^*;\mathcal{G}_{u,v};\bal), \label{apx-se}
\end{align}
For suugraph encoding function searching in Eq.~(\ref{apx-se}), 
following~\cite{zhang2020interstellar},
we adopt stochastic relaxation on $\bal$ and natural policy gradient strategy~\cite{akimoto2019adaptive} to obtain the optimal subgraph encoding function $\bal^*$.
For subgraph selection in Eq.~(\ref{apx-ss}),
we obtain the optimal subgraph $\mathcal{G}_{u,v}^*$ by preserving the subgraph with the largest probability $p_{u,v}^{i,j}$, i.e., 
\begin{align}
	\mathbf{z}_{u,v}^{i,j} &= f(\mathcal{G}_{u,v}^{i,j}),\\
	\beta_{u,v}^{i,j} &= g(\mathbf{z}_{u,v}^{i,j}),\\
	p_{u,v}^{i,j} &= \frac{\exp(\log (\beta_{u,v}^{i,j} + \textbf{G}_{i,j})/\tau)}{\sum_{i',j'=1}^\eta \exp(\log (\beta_{u,v}^{i',j'} + \textbf{G}_{i',j'})/\tau)},\\
	\mathcal{G}_{u,v}^*&=\argmax\limits_{\mathcal{G}^{i,j}_{u,v}}p_{u,v}^{i,j}( \mathcal{G}^{i,j}_{u,v}\in{S}_{u,v}).
\end{align}

\section{More Experiment Setting}
\subsection{Datasets}\label{apd-dataset}

Experiments are performed on two public benchmark DDI datasets: DrugBank and TWOSIDES. 
\paragraph{DrugBank}
DrugBank dataset contains 1,710 drugs and drug pairs, 
which are related to 86 types of pharmacological interactions between drugs, 
such as increase of anticoagulant activity, decrease of excretion rate and etc.
\paragraph{TWOSIDES}
TWOSIDES dataset contains 604 drugs and drug pairs with 200 drug side effects as interaction labels. For each edge, it may be associated with multiple interactions.  

The detailed descriptions for datasets are presented in Table \ref{tab:statics} and Table~\ref{tab:relation_statics}.
\begin{table*}[h]
	\centering
	\caption{The statistics of the datasets.}
	\setlength\tabcolsep{20pt}
	\begin{tabular}{cccc}
		\toprule
		Dataset & \#nodes & \#edges & \#interaction types \\
		\midrule
		DrugBank & 1,710 & 134641 & 86 \\
		TWOSIDES & 604 & 57778 & 200 \\
		\bottomrule
	\end{tabular}
	\label{tab:statics}
	\vspace{-10px}
\end{table*}

\begin{table*}[h]
	\centering
	\caption{Diverse semantic properties in drug-drug interactions.}
	\begin{tabular}{cccc}
		\toprule
		Dataset & Interaction Type  & Examples & Semantic Property \\
		\midrule
		DrugBank & Metabolic levels-based  & \makecell{\#Drug1 may decrease the \\ excretion rate of \#Drug2}  & \makecell{asymmetry \\$(r(x,y)\nRightarrow r(y,x))$ } \\
		TWOSIDES & Phenotype-based & \makecell{Combination of \#Drug 1 and \\ \#Drug 2 may cause headaches} & \makecell{symmetry \\ $(r(x,y)\Rightarrow r(y,x))$} \\
		\bottomrule
	\end{tabular}
	\label{tab:relation_statics}
	\vspace{-10px}
\end{table*}

\subsection{Evaluation Metric}\label{apd-eva}

We follow~\cite{yu2021sumgnn} to evaluate our method. 
Specifically, 
in terms of the multi-class prediction on DrugBank, 
we followc\cite{yu2021sumgnn} and evaluate the performance by three metrics: 
(i) Macro F1 score~(Macro F1) is computed by taking the arithmetic mean (aka unweighted mean) of all the per-class F1 scores. 
(ii) Accuracy~(ACC) is calculated by dividing the number of correct predictions by the total prediction number. 
(iii) Coken's Kappa~(Cohen's $\kappa$) measures inter-rater reliability. 
As to the multi-label prediction on TWOSIDES, 
we consider the following measure and use the average performance over all interaction types: 
(i) ROC-AUC~(AUROC) stands for “Area Under the Curve~(AUC)” of the “Receiver Operating Characteristic~(ROC)” curve. 
(ii) PR-AUC~(AUPRC) is the average area under precision-recall curve. 
(iii) AP@50 is the average precision at 50.

\subsection{Implementation and Hyperparameters}\label{apd-implement}

All the experiments are implemented in Python with the PyTorch framework~\cite{paszke2019pytorch} and run on a server machine with single NVIDIA RTX 3090 GPU with 24GB memory and 64GB of RAM.
Our code is added in the supplementary material.

For CSSE-DDI,
we set the epoch to 400 for training supernet and set the epoch to 400 for training sub-supernets.
We set the the temperature parameter as 0.05.
Repeat 5 times with different seeds,
we can get 5 candidates.
The searched candidates are finetuned individually with the hyper-parameters.
In the stage of fine-tuning, we use the ReduceLROnPlateau scheduler to adjust the learning rate dynamically. 
Each candidate has 10 hyper steps. 
In each hyper step, 
a set of hyperparameter will be sampled from Table~\ref{tab-hyper-finetune}.
\begin{table}[h]
	\centering
	\caption{Hyperparameters we used during the fine-tuning stage.}
	\begin{tabular}{cc}
		\toprule
		\textbf{Hyperparameter} & \textbf{Value range}              \\ 
		\midrule
		Learning rate & $[10^{-3.1},10^{-2.9}]$ \\
		\midrule
		Weight decay   &           $[10^{-5},10^{-3}]$         \\ 
		\bottomrule
	\end{tabular}
	\label{tab-hyper-finetune}
\end{table}

\section{More Experimental Results}
\subsection{Subgraph Scope Distribution Analysis}

\begin{figure}[h]
	\centering
	\includegraphics[width=0.4\linewidth]{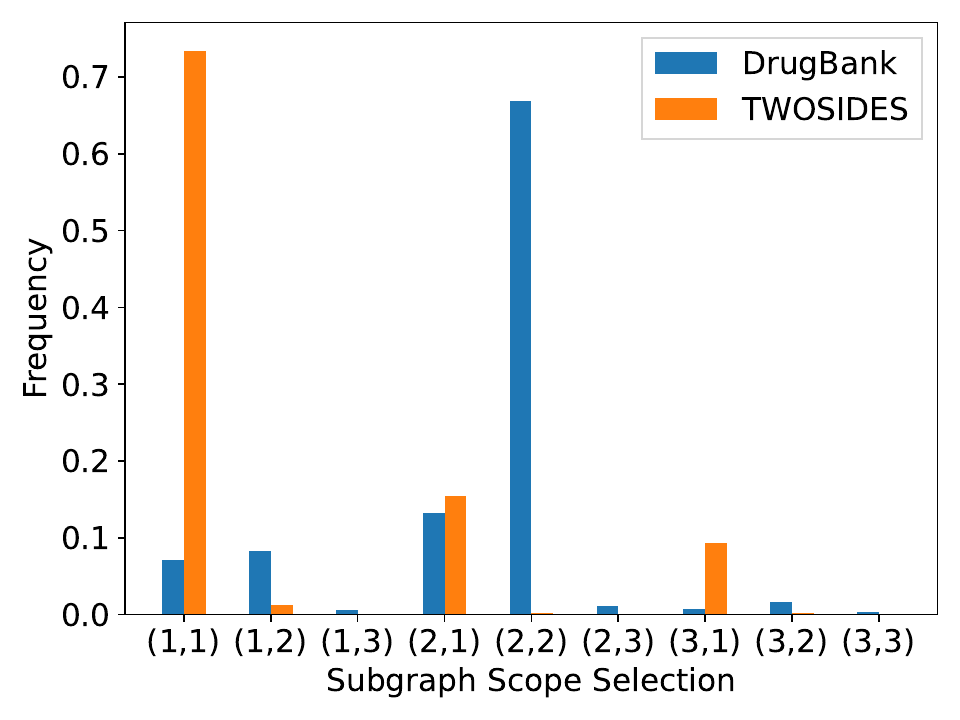}
	\caption{Distribution of the searched subgraph scopes by CSSE-DDI on all benchmark datasets.}
	\label{fig-distr}
\end{figure}

We visualize the learned distributions of subgraph scope on all datasets by using CSSE-DDI in Figure~\ref{fig-distr}.
By comparing the distributions across different benchmarks,
we have the following observation:
CSSE-DDI can effectively learn different subgraph scope distributions for various datasets.
By identifing specific subgraph scopes for different queries,
CSSE-DDI is able to precisely control the extent of information propagation required for reasoning about the interactions of different drug pairs.
In addition, 
our method can skip some subgraph scopes if they are not optimal for any queries.
For example,
no queries are assigned to the propagation scope $(3,3)$ on TWOSIDES dataset.
It is worth mentioning that our searched subgraph scopes are consistent with the sensitivity analysis results for the hop of subgraph in SumGNN~\cite{yu2021sumgnn}, 
which further validates the effectiveness of our approach.

\section{Some Discussions about Checklist}
\subsection{Limitations}\label{apx-limit}
There are three limitations for CSSE-DDI. 
(1) CSSE-DDI is focused on method design rather than system design.
In the future, we will co-design the algorithm and
the system to further improve the efficiency. 
(2) At present, CSSE-DDI only search for data-specific components of subgraph-based pipeline, 
while hyper-parameters are also important
for DDI prediction. 
A promising direction is to explore
how to efficiently search network architectures and
hyper-parameters simultaneously.


\newpage
\section*{NeurIPS Paper Checklist}

The checklist is designed to encourage best practices for responsible machine learning research, addressing issues of reproducibility, transparency, research ethics, and societal impact. Do not remove the checklist: {\bf The papers not including the checklist will be desk rejected.} The checklist should follow the references and follow the (optional) supplemental material.  The checklist does NOT count towards the page
limit. 

Please read the checklist guidelines carefully for information on how to answer these questions. For each question in the checklist:
\begin{itemize}
    \item You should answer \answerYes{}, \answerNo{}, or \answerNA{}.
    \item \answerNA{} means either that the question is Not Applicable for that particular paper or the relevant information is Not Available.
    \item Please provide a short (1–2 sentence) justification right after your answer (even for NA). 
\end{itemize}

{\bf The checklist answers are an integral part of your paper submission.} They are visible to the reviewers, area chairs, senior area chairs, and ethics reviewers. You will be asked to also include it (after eventual revisions) with the final version of your paper, and its final version will be published with the paper.

The reviewers of your paper will be asked to use the checklist as one of the factors in their evaluation. While "\answerYes{}" is generally preferable to "\answerNo{}", it is perfectly acceptable to answer "\answerNo{}" provided a proper justification is given (e.g., "error bars are not reported because it would be too computationally expensive" or "we were unable to find the license for the dataset we used"). In general, answering "\answerNo{}" or "\answerNA{}" is not grounds for rejection. While the questions are phrased in a binary way, we acknowledge that the true answer is often more nuanced, so please just use your best judgment and write a justification to elaborate. All supporting evidence can appear either in the main paper or the supplemental material, provided in appendix. If you answer \answerYes{} to a question, in the justification please point to the section(s) where related material for the question can be found.

IMPORTANT, please:
\begin{itemize}
    \item {\bf Delete this instruction block, but keep the section heading ``NeurIPS paper checklist"},
    \item  {\bf Keep the checklist subsection headings, questions/answers and guidelines below.}
    \item {\bf Do not modify the questions and only use the provided macros for your answers}.
\end{itemize}


\begin{enumerate}

\item {\bf Claims}
    \item[] Question: Do the main claims made in the abstract and introduction accurately reflect the paper's contributions and scope?
    \item[] Answer: \answerYes{} 
    \item[] Justification: The abstract and introduction clearly state the claims made, including the contributions made in our paper and important assumptions and limitations.
    \item[] Guidelines:
    \begin{itemize}
        \item The answer NA means that the abstract and introduction do not include the claims made in the paper.
        \item The abstract and/or introduction should clearly state the claims made, including the contributions made in the paper and important assumptions and limitations. A No or NA answer to this question will not be perceived well by the reviewers. 
        \item The claims made should match theoretical and experimental results, and reflect how much the results can be expected to generalize to other settings. 
        \item It is fine to include aspirational goals as motivation as long as it is clear that these goals are not attained by the paper. 
    \end{itemize}

\item {\bf Limitations}
    \item[] Question: Does the paper discuss the limitations of the work performed by the authors?
    \item[] Answer: \answerYes{} 
    \item[] Justification: We discuss the limitations in Section~\ref{apx-limit} of the Appendix.
    \item[] Guidelines:
    \begin{itemize}
        \item The answer NA means that the paper has no limitation while the answer No means that the paper has limitations, but those are not discussed in the paper. 
        \item The authors are encouraged to create a separate "Limitations" section in their paper.
        \item The paper should point out any strong assumptions and how robust the results are to violations of these assumptions (e.g., independence assumptions, noiseless settings, model well-specification, asymptotic approximations only holding locally). The authors should reflect on how these assumptions might be violated in practice and what the implications would be.
        \item The authors should reflect on the scope of the claims made, e.g., if the approach was only tested on a few datasets or with a few runs. In general, empirical results often depend on implicit assumptions, which should be articulated.
        \item The authors should reflect on the factors that influence the performance of the approach. For example, a facial recognition algorithm may perform poorly when image resolution is low or images are taken in low lighting. Or a speech-to-text system might not be used reliably to provide closed captions for online lectures because it fails to handle technical jargon.
        \item The authors should discuss the computational efficiency of the proposed algorithms and how they scale with dataset size.
        \item If applicable, the authors should discuss possible limitations of their approach to address problems of privacy and fairness.
        \item While the authors might fear that complete honesty about limitations might be used by reviewers as grounds for rejection, a worse outcome might be that reviewers discover limitations that aren't acknowledged in the paper. The authors should use their best judgment and recognize that individual actions in favor of transparency play an important role in developing norms that preserve the integrity of the community. Reviewers will be specifically instructed to not penalize honesty concerning limitations.
    \end{itemize}

\item {\bf Theory Assumptions and Proofs}
    \item[] Question: For each theoretical result, does the paper provide the full set of assumptions and a complete (and correct) proof?
    \item[] Answer: \answerNA{} 
    \item[] Justification: Our paper does not include theoretical results.
    \item[] Guidelines:
    \begin{itemize}
        \item The answer NA means that the paper does not include theoretical results. 
        \item All the theorems, formulas, and proofs in the paper should be numbered and cross-referenced.
        \item All assumptions should be clearly stated or referenced in the statement of any theorems.
        \item The proofs can either appear in the main paper or the supplemental material, but if they appear in the supplemental material, the authors are encouraged to provide a short proof sketch to provide intuition. 
        \item Inversely, any informal proof provided in the core of the paper should be complemented by formal proofs provided in appendix or supplemental material.
        \item Theorems and Lemmas that the proof relies upon should be properly referenced. 
    \end{itemize}

    \item {\bf Experimental Result Reproducibility}
    \item[] Question: Does the paper fully disclose all the information needed to reproduce the main experimental results of the paper to the extent that it affects the main claims and/or conclusions of the paper (regardless of whether the code and data are provided or not)?
    \item[] Answer: \answerYes{} 
    \item[] Justification: We provide the complete code that runs correctly and hyperparameter configurations in the supplemental material and Section~\ref{apd-implement} to ensure reproducibility and transparency.
    \item[] Guidelines:
    \begin{itemize}
        \item The answer NA means that the paper does not include experiments.
        \item If the paper includes experiments, a No answer to this question will not be perceived well by the reviewers: Making the paper reproducible is important, regardless of whether the code and data are provided or not.
        \item If the contribution is a dataset and/or model, the authors should describe the steps taken to make their results reproducible or verifiable. 
        \item Depending on the contribution, reproducibility can be accomplished in various ways. For example, if the contribution is a novel architecture, describing the architecture fully might suffice, or if the contribution is a specific model and empirical evaluation, it may be necessary to either make it possible for others to replicate the model with the same dataset, or provide access to the model. In general. releasing code and data is often one good way to accomplish this, but reproducibility can also be provided via detailed instructions for how to replicate the results, access to a hosted model (e.g., in the case of a large language model), releasing of a model checkpoint, or other means that are appropriate to the research performed.
        \item While NeurIPS does not require releasing code, the conference does require all submissions to provide some reasonable avenue for reproducibility, which may depend on the nature of the contribution. For example
        \begin{enumerate}
            \item If the contribution is primarily a new algorithm, the paper should make it clear how to reproduce that algorithm.
            \item If the contribution is primarily a new model architecture, the paper should describe the architecture clearly and fully.
            \item If the contribution is a new model (e.g., a large language model), then there should either be a way to access this model for reproducing the results or a way to reproduce the model (e.g., with an open-source dataset or instructions for how to construct the dataset).
            \item We recognize that reproducibility may be tricky in some cases, in which case authors are welcome to describe the particular way they provide for reproducibility. In the case of closed-source models, it may be that access to the model is limited in some way (e.g., to registered users), but it should be possible for other researchers to have some path to reproducing or verifying the results.
        \end{enumerate}
    \end{itemize}

\item {\bf Open access to data and code}
    \item[] Question: Does the paper provide open access to the data and code, with sufficient instructions to faithfully reproduce the main experimental results, as described in supplemental material?
    \item[] Answer: \answerYes{} 
    \item[] Justification: We provide the datasets, the complete code that runs correctly and hyperparameter configurations in the supplemental material and Section~\ref{apd-implement} to ensure reproducibility and transparency.
    \item[] Guidelines:
    \begin{itemize}
        \item The answer NA means that paper does not include experiments requiring code.
        \item Please see the NeurIPS code and data submission guidelines (\url{https://nips.cc/public/guides/CodeSubmissionPolicy}) for more details.
        \item While we encourage the release of code and data, we understand that this might not be possible, so “No” is an acceptable answer. Papers cannot be rejected simply for not including code, unless this is central to the contribution (e.g., for a new open-source benchmark).
        \item The instructions should contain the exact command and environment needed to run to reproduce the results. See the NeurIPS code and data submission guidelines (\url{https://nips.cc/public/guides/CodeSubmissionPolicy}) for more details.
        \item The authors should provide instructions on data access and preparation, including how to access the raw data, preprocessed data, intermediate data, and generated data, etc.
        \item The authors should provide scripts to reproduce all experimental results for the new proposed method and baselines. If only a subset of experiments are reproducible, they should state which ones are omitted from the script and why.
        \item At submission time, to preserve anonymity, the authors should release anonymized versions (if applicable).
        \item Providing as much information as possible in supplemental material (appended to the paper) is recommended, but including URLs to data and code is permitted.
    \end{itemize}

\item {\bf Experimental Setting/Details}
    \item[] Question: Does the paper specify all the training and test details (e.g., data splits, hyperparameters, how they were chosen, type of optimizer, etc.) necessary to understand the results?
    \item[] Answer: \answerYes{} 
    \item[] Justification: We provide the data splits, hyperparameter configurations, and other experimental details in the supplemental material and Section~\ref{apd-implement} to ensure reproducibility and transparency.
    \item[] Guidelines:
    \begin{itemize}
        \item The answer NA means that the paper does not include experiments.
        \item The experimental setting should be presented in the core of the paper to a level of detail that is necessary to appreciate the results and make sense of them.
        \item The full details can be provided either with the code, in appendix, or as supplemental material.
    \end{itemize}

\item {\bf Experiment Statistical Significance}
    \item[] Question: Does the paper report error bars suitably and correctly defined or other appropriate information about the statistical significance of the experiments?
    \item[] Answer: \answerYes{} 
    \item[] Justification: All of the methods are run for five times on the different random seeds with mean value and
    standard deviation reported on the testing data, as shown in Table~\ref{tab:main}. 
    \item[] Guidelines:
    \begin{itemize}
        \item The answer NA means that the paper does not include experiments.
        \item The authors should answer "Yes" if the results are accompanied by error bars, confidence intervals, or statistical significance tests, at least for the experiments that support the main claims of the paper.
        \item The factors of variability that the error bars are capturing should be clearly stated (for example, train/test split, initialization, random drawing of some parameter, or overall run with given experimental conditions).
        \item The method for calculating the error bars should be explained (closed form formula, call to a library function, bootstrap, etc.)
        \item The assumptions made should be given (e.g., Normally distributed errors).
        \item It should be clear whether the error bar is the standard deviation or the standard error of the mean.
        \item It is OK to report 1-sigma error bars, but one should state it. The authors should preferably report a 2-sigma error bar than state that they have a 96\% CI, if the hypothesis of Normality of errors is not verified.
        \item For asymmetric distributions, the authors should be careful not to show in tables or figures symmetric error bars that would yield results that are out of range (e.g. negative error rates).
        \item If error bars are reported in tables or plots, The authors should explain in the text how they were calculated and reference the corresponding figures or tables in the text.
    \end{itemize}

\item {\bf Experiments Compute Resources}
    \item[] Question: For each experiment, does the paper provide sufficient information on the computer resources (type of compute workers, memory, time of execution) needed to reproduce the experiments?
    \item[] Answer: \answerYes{} 
    \item[] Justification:  We provide the configuration of running environment in the supplemental material and Section~\ref{apd-implement} to ensure reproducibility and transparency.
    \item[] Guidelines:
    \begin{itemize}
        \item The answer NA means that the paper does not include experiments.
        \item The paper should indicate the type of compute workers CPU or GPU, internal cluster, or cloud provider, including relevant memory and storage.
        \item The paper should provide the amount of compute required for each of the individual experimental runs as well as estimate the total compute. 
        \item The paper should disclose whether the full research project required more compute than the experiments reported in the paper (e.g., preliminary or failed experiments that didn't make it into the paper). 
    \end{itemize}
    
\item {\bf Code Of Ethics}
    \item[] Question: Does the research conducted in the paper conform, in every respect, with the NeurIPS Code of Ethics \url{https://neurips.cc/public/EthicsGuidelines}?
    \item[] Answer: \answerYes{} 
    \item[] Justification: We would claim that this work does not raise any ethical concerns. Besides, this work does not involve any human subjects, practices to data set releases, potentially harmful insights, methodologies
    and applications, potential conflicts of interest and sponsorship, discrimination/bias/fairness concerns, privacy and security issues, legal compliance, and research integrity issues.
    \item[] Guidelines:
    \begin{itemize}
        \item The answer NA means that the authors have not reviewed the NeurIPS Code of Ethics.
        \item If the authors answer No, they should explain the special circumstances that require a deviation from the Code of Ethics.
        \item The authors should make sure to preserve anonymity (e.g., if there is a special consideration due to laws or regulations in their jurisdiction).
    \end{itemize}

\item {\bf Broader Impacts}
    \item[] Question: Does the paper discuss both potential positive societal impacts and negative societal impacts of the work performed?
    \item[] Answer: \answerYes{} 
    \item[] Justification: We believe that this work is expected to have a positive impact in the field of health care and medicine, and by predicting drug-drug interactions, the method has a positive effect in reducing experimental costs and assisting in the prediction of drug-drug interactions.
    \item[] Guidelines:
    \begin{itemize}
        \item The answer NA means that there is no societal impact of the work performed.
        \item If the authors answer NA or No, they should explain why their work has no societal impact or why the paper does not address societal impact.
        \item Examples of negative societal impacts include potential malicious or unintended uses (e.g., disinformation, generating fake profiles, surveillance), fairness considerations (e.g., deployment of technologies that could make decisions that unfairly impact specific groups), privacy considerations, and security considerations.
        \item The conference expects that many papers will be foundational research and not tied to particular applications, let alone deployments. However, if there is a direct path to any negative applications, the authors should point it out. For example, it is legitimate to point out that an improvement in the quality of generative models could be used to generate deepfakes for disinformation. On the other hand, it is not needed to point out that a generic algorithm for optimizing neural networks could enable people to train models that generate Deepfakes faster.
        \item The authors should consider possible harms that could arise when the technology is being used as intended and functioning correctly, harms that could arise when the technology is being used as intended but gives incorrect results, and harms following from (intentional or unintentional) misuse of the technology.
        \item If there are negative societal impacts, the authors could also discuss possible mitigation strategies (e.g., gated release of models, providing defenses in addition to attacks, mechanisms for monitoring misuse, mechanisms to monitor how a system learns from feedback over time, improving the efficiency and accessibility of ML).
    \end{itemize}
    
\item {\bf Safeguards}
    \item[] Question: Does the paper describe safeguards that have been put in place for responsible release of data or models that have a high risk for misuse (e.g., pretrained language models, image generators, or scraped datasets)?
    \item[] Answer: \answerNA{} 
    \item[] Justification: This paper poses no such risks.
    \item[] Guidelines:
    \begin{itemize}
        \item The answer NA means that the paper poses no such risks.
        \item Released models that have a high risk for misuse or dual-use should be released with necessary safeguards to allow for controlled use of the model, for example by requiring that users adhere to usage guidelines or restrictions to access the model or implementing safety filters. 
        \item Datasets that have been scraped from the Internet could pose safety risks. The authors should describe how they avoided releasing unsafe images.
        \item We recognize that providing effective safeguards is challenging, and many papers do not require this, but we encourage authors to take this into account and make a best faith effort.
    \end{itemize}

\item {\bf Licenses for existing assets}
    \item[] Question: Are the creators or original owners of assets (e.g., code, data, models), used in the paper, properly credited and are the license and terms of use explicitly mentioned and properly respected?
    \item[] Answer: \answerYes{} 
    \item[] Justification: We cite the original paper that produced the code package or dataset.
    \item[] Guidelines:
    \begin{itemize}
        \item The answer NA means that the paper does not use existing assets.
        \item The authors should cite the original paper that produced the code package or dataset.
        \item The authors should state which version of the asset is used and, if possible, include a URL.
        \item The name of the license (e.g., CC-BY 4.0) should be included for each asset.
        \item For scraped data from a particular source (e.g., website), the copyright and terms of service of that source should be provided.
        \item If assets are released, the license, copyright information, and terms of use in the package should be provided. For popular datasets, \url{paperswithcode.com/datasets} has curated licenses for some datasets. Their licensing guide can help determine the license of a dataset.
        \item For existing datasets that are re-packaged, both the original license and the license of the derived asset (if it has changed) should be provided.
        \item If this information is not available online, the authors are encouraged to reach out to the asset's creators.
    \end{itemize}

\item {\bf New Assets}
    \item[] Question: Are new assets introduced in the paper well documented and is the documentation provided alongside the assets?
    \item[] Answer: \answerYes{} 
    \item[] Justification: The assets we submitted have detailed documentation.
    \item[] Guidelines:
    \begin{itemize}
        \item The answer NA means that the paper does not release new assets.
        \item Researchers should communicate the details of the dataset/code/model as part of their submissions via structured templates. This includes details about training, license, limitations, etc. 
        \item The paper should discuss whether and how consent was obtained from people whose asset is used.
        \item At submission time, remember to anonymize your assets (if applicable). You can either create an anonymized URL or include an anonymized zip file.
    \end{itemize}

\item {\bf Crowdsourcing and Research with Human Subjects}
    \item[] Question: For crowdsourcing experiments and research with human subjects, does the paper include the full text of instructions given to participants and screenshots, if applicable, as well as details about compensation (if any)? 
    \item[] Answer: \answerNA{} 
    \item[] Justification: The paper does not involve crowdsourcing nor research with human subjects.
    \item[] Guidelines:
    \begin{itemize}
        \item The answer NA means that the paper does not involve crowdsourcing nor research with human subjects.
        \item Including this information in the supplemental material is fine, but if the main contribution of the paper involves human subjects, then as much detail as possible should be included in the main paper. 
        \item According to the NeurIPS Code of Ethics, workers involved in data collection, curation, or other labor should be paid at least the minimum wage in the country of the data collector. 
    \end{itemize}

\item {\bf Institutional Review Board (IRB) Approvals or Equivalent for Research with Human Subjects}
    \item[] Question: Does the paper describe potential risks incurred by study participants, whether such risks were disclosed to the subjects, and whether Institutional Review Board (IRB) approvals (or an equivalent approval/review based on the requirements of your country or institution) were obtained?
    \item[] Answer: \answerNA{} 
    \item[] Justification: The paper does not involve crowdsourcing nor research with human subjects.
    \item[] Guidelines:
    \begin{itemize}
        \item The answer NA means that the paper does not involve crowdsourcing nor research with human subjects.
        \item Depending on the country in which research is conducted, IRB approval (or equivalent) may be required for any human subjects research. If you obtained IRB approval, you should clearly state this in the paper. 
        \item We recognize that the procedures for this may vary significantly between institutions and locations, and we expect authors to adhere to the NeurIPS Code of Ethics and the guidelines for their institution. 
        \item For initial submissions, do not include any information that would break anonymity (if applicable), such as the institution conducting the review.
    \end{itemize}

\end{enumerate}

\end{document}